\documentclass[runningheads]{llncs}
\usepackage[T1]{fontenc}
\usepackage{amsmath} %
\usepackage{subcaption} %
\usepackage{graphicx}

\usepackage[numbers,sort&compress,sectionbib]{natbib}

\makeatletter \renewcommand\@biblabel[1]{#1.} \makeatother %

\begin{document}
\title{EdgeAL: An Edge Estimation Based Active Learning Approach for OCT Segmentation}
\titlerunning{EdgeAL}
\author{Md Abdul Kadir\inst{1}\orcidID{0000-0002-8420-2536} \and
Hasan Md Tusfiqur Alam \inst{1}\orcidID{0000-0003-1479-7690} \and
Daniel Sonntag\inst{1,2}\orcidID{0000-0002-8857-8709}}
\authorrunning{M. A. Kadir et al.}

\institute{
German Research Center for Artificial Intelligence (DFKI), Germany \and
University of Oldenburg, Oldenburg, Germany\\
\email{\{abdul.kadir, hasan.alam, daniel.sonntag\}@dfki.de}
}

\maketitle              %
\begin{abstract}
Active learning algorithms have become increasingly popular for training models with limited data. However, selecting data for annotation remains a challenging problem due to the limited information available on unseen data. To address this issue, we propose EdgeAL, which utilizes the edge information of unseen images as {\it a priori} information for measuring uncertainty. The uncertainty is quantified by analyzing the divergence and entropy in model predictions across edges. This measure is then used to select superpixels for annotation. We demonstrate the effectiveness of EdgeAL on multi-class Optical Coherence Tomography (OCT) segmentation tasks, where we achieved a 99\% dice score while reducing the annotation label cost to 12\%, 2.3\%, and 3\%, respectively, on three publicly available datasets (Duke, AROI, and UMN).  The source code is available at \url{https://github.com/Mak-Ta-Reque/EdgeAL}

\end{abstract}
\section{Introduction}

In recent years, Deep Learning (DL) based methods have achieved considerable success in the medical domain for tasks including disease diagnosis and clinical feature segmentation \cite{yuan2022multiscale,nguyen2020visually}. However, their progress is often constrained as they require large labelled datasets. Labelling medical image data is a labour-intensive and time-consuming process that needs the careful attention of clinical experts. Active learning (AL) can benefit the iterative improvement of any intelligent diagnosis system by reducing the burden of extensive annotation effort \cite{nath2022warm,alam2023drg}.

Ophthalmologists use the segmentation of ocular Optical Coherence Tomography (OCT) images to diagnose, and treatment of eye diseases such as Diabetic Retinopathy (DR) and Diabetic Macular Edema (DME) \cite{farshad2022net}. Here, we propose a novel Edge estimation-based Active Learning EdgeAL framework for OCT image segmentation that leverages prediction uncertainty across the boundaries of the semantic regions of input images. The Edge information is one of the image's most salient features, and it can boost segmentation accuracy when integrated into neural model training \cite{lu2023multi}. We formulate a novel acquisition function that leverages the variance of the predicted score across the gradient surface of the input to measure uncertainty. Empirical results show that EdgeAL achieves \textit{state-of-the-art} performance with minimal annotation samples, using a seed set as small as 2\% of unlabeled data.

\section{Related Work}

Active learning is a cost-effective strategy that selects the most informative samples for annotation to improve model performance based on uncertainty \cite{lee2018robust}, data distribution \cite{samrath2019variational}, expected model change \cite{dai2020suggestive}, and other criteria \cite{bai2022discrepancy}. A simpler way to measure uncertainty can be realized using posterior probabilities of predictions, such as selecting instances with the least confidence \cite{lee2018robust,joshi2009multi}, or computing class entropy \cite{luo2013latent}.

Some uncertainty-based approaches have been directly used with deep neural networks \cite{siddiqui2020viewal}. \citet{gal2017deep} propose dropout-base Monte Carlo (MC) sampling to obtain uncertainty estimation. It uses multiple forward passes with dropout at different layers to generate uncertainty during inference.
Ensemble-based methods also have been widely used where the variance between the prediction outcomes from a collection of models serve as the uncertainty \cite{nath2020diminishing,yang2017suggestive,sener2017active}. 

Many AL methods have been adopted for segmentation tasks \cite{nath2020diminishing,gorriz2017cost,mackowiak2018cereals}. 
\citet{gorriz2017cost} propose an AL framework Melanoma segmentation by extending Cost-Effective Active Learning (CEAL) \cite{wang2016cost} algorithm where complimentary samples of both high and low confidence are selected for annotation. \citet{mackowiak2018cereals} use a region-based selection approach and estimate model uncertainty using MC dropout to reduce human-annotation cost. \citet{nath2020diminishing} propose an ensemble-based method where multiple AL frameworks are jointly optimized, and a \textit{query-by-committee} approach is adopted for sample selection. These methods do not consider any prior information to estimate uncertainty. Authors in \cite{siddiqui2020viewal} propose an AL framework for multi-view datasets \cite{muslea2006active} segmentation task where model uncertainty is estimated based on \textit{Kullback-Leibler} (KL) divergence of posterior probability distributions for a disjoint subset of prior features such as depth, and camera position.

However, viewpoint information is not always available in medical imaging. We leverage edge information as a prior for AL sampling based on previous studies where edge information has improved the performance of segmentation tasks \cite{lu2023multi}. To our knowledge, there has yet to be any exploration of using image edges as an \textit{a priori} in active learning.

There has not been sufficient work other than \cite{li2021unsupervised} related to Active Learning for OCT segmentation. Their approach requires foundation models \cite{khan2020survey} to be pre-trained on large-scale datasets in similar domains, which could be infeasible to collect due to data privacy. On the other hand, our method requires a few samples ($\sim$2\%) for initial training, overcoming the limitation of the need for a large dataset.

\section{Methodology}
\begin{figure}
\includegraphics[width=\textwidth, trim={3mm .5mm 0mm 1mm}, clip]{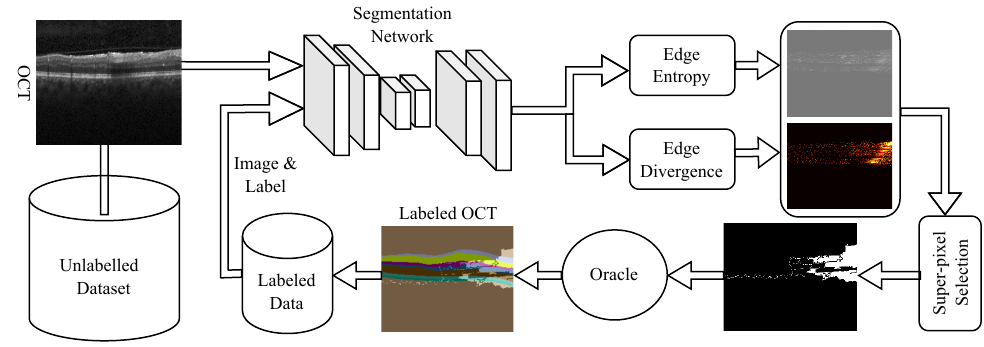}
\caption{The figure above illustrates the workflow of our AL framework. It first computes an OCT image's \textit{edge entropy} and \textit{edge divergence} maps. Later, it calculates the overlaps between superpixels based on the divergence and entropy map to recommend an annotation region.} 
\label{fig1}
\end{figure}
Figure \ref{fig1} shows that our active learning technique consists of four major stages. First, we train the network on a subset of labeled images, usually a tiny percentage of the total collection (e.g., 2\%). Following that, we compute uncertainty values for input instances and input areas. Based on this knowledge, we select superpixels to label and obtain annotations from a simulated oracle.
\label{sec3_1}
\subsection{Segmentation network}
We trained our OCT semantic segmentation model using a randomly selected small portion of the labeled data $D_s$, seed set, keeping the rest for oracle imitation. We choose Y-net-gen-ffc (YN\textsuperscript{*}) without pre-retrained weight initialization as our primary architecture due to its superior performance \cite{farshad2022net}.%

\subsection{Uncertainty in prediction}
EdgeAL seeks to improve the model's performance by querying uncertain areas on unlabeled data $D_u$ after training it on a seed set $D_s$. 
To accomplish this, we have created a novel edge-based uncertainty measurement method. We compute the \textit{edge entropy} score and \textit{edge divergence} score - to assess the prediction ambiguity associated with the edges. Figure \ref{fig:four_images} depicts examples of input OCT, measured \textit{edge entropy}, and \textit{edge kl-divergence} corresponding to the input.

\subsubsection{Entropy Score on Edges}
Analyzing the edges of raw OCT inputs yields critical information on features and texture in images. They may look noisy, but they summarize all the alterations in a picture. The Sobel operator can be used to identify edges in the input image \cite{lu2023multi}. Let us define the normalized absolute value of edges of an image $I_i$ by $S_i$. $|\nabla I_i|$ is the absolute gradient.

\begin{equation*}
S_i = \frac {|\nabla I_i| - min(|\nabla I_i|)}{max(|\nabla I_i|) - min(|\nabla I_i|)}
\end{equation*}

To determine the probability that each pixel in an image belongs to a particular class $c$, we use the output of our network, denoted as $P_{i}^{(m,n)}(c)$. We adopt Monte Carlo (MC) dropout simulation for uncertainty sampling and average predictions over $\vert D \vert$ occurrence from \cite{gal2017deep}. Consequently, an MC probability distribution depicts the chance of a pixel at location $(m,n)$ in picture $I_i$ belonging to a class $c$, and $C$ is the set of segmentation classes. We run MC dropouts $\vert D \vert$ times during the neural network assessment mode and measure $P_{i}^{(m,n)}(c)$ using Equation \ref{eq_1}.

\begin{equation}
P_{i}^{(m,n)}(c) = \frac{1}{\vert D \vert} \sum_{d=1}^D P_{i,d}^{(m,n)}(c)
\label{eq_1}
\end{equation}

Following \citet{zhao2021_calibrate}, we apply contextual calibration on $P_{i}^{(m,n)}(c)$ by $S_i$ to prioritize significant input surface variations. Now, $S_i$ is linked with a probability distribution, with $\phi_{i}^{(m,n)}(c)$  having information about the edges of input. This formulation makes our implementation unique from other active learning methods in image segmentation.

\begin{equation}
\phi_{i}^{m,n} (c)=  \frac {e^ {P_{i}^{(m,n)}(c) \bullet   S_i {(m,n)}}}{{\sum_{k \in C} e^{ P_{i}^{(m,n)} (k) \bullet   S_i^{(m,n)}}}}
\label{eq_2}
\end{equation}

We name $\phi_{i}^{m,n} (c)$ as contextual probability and define our \textit{edge entropy} by following entropy formula of \cite{luo2013latent}.

\begin{equation}
    EE_{i}^{m,n} = - \sum_{c \in C} \phi_{i}^{m,n} (c) \log ( \phi_{i}^{m,n} (c))
\end{equation}

\subsubsection{Divergence Score on Edges}
In areas with strong edges/gradients, \textit{edge entropy} reflects the degree of inconsistency in the network's prediction for each input pixel. However, the degree of this uncertainty must also be measured. \textit{KL-divergence} is used to measure the difference in inconsistency between $P_{i}^{(m,n)}$ and $\phi_{i}^{(m,n)}$ for a pixel $(m, n)$ in an input image based on the idea of self-knowledge distillation $I_i$ \cite{Yun_2020_CVPR}. The \textit{edge divergence} $ED_{i}^{m,n}$ score can be formalized using equation \ref{eq_1} and \ref{eq_2}.
\begin{equation*} 
    ED_{i}^{(m,n)} = D_{KL} \big ( P_{i}^{(m,n)} || \phi_{i}^{(m,n)} \big )
\end{equation*}

\begin{figure}[ht!]
\centering
\begin{subfigure}[t]{0.23\textwidth}
\centering
\includegraphics[width=\textwidth, trim=0mm 10mm 0mm 10mm, clip]{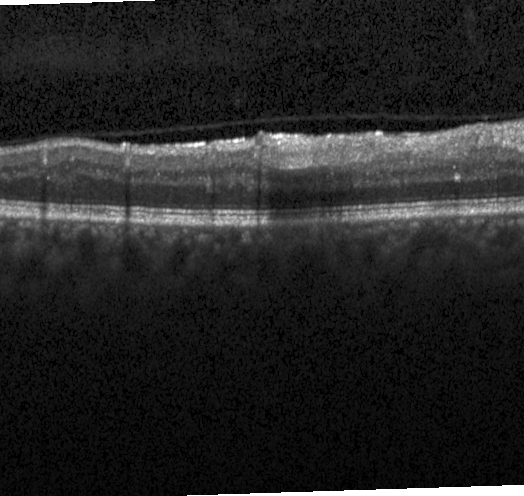}
\caption{OCT}
\label{fig:subfig1}
\end{subfigure}%
\hfill
\begin{subfigure}[t]{0.23\textwidth}
\centering
\includegraphics[width=\textwidth, trim=0mm 5mm 0mm 7mm, clip]{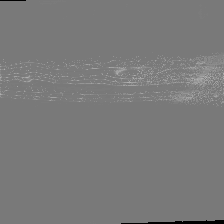}
\caption{Entropy map}
\label{fig:subfig2}
\end{subfigure}%
\hfill
\begin{subfigure}[t]{0.23\textwidth}
\centering
\includegraphics[width=\textwidth, trim=0mm 5mm 0mm 3mm, clip]{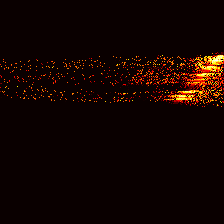}
\caption{Divergence map}
\label{fig:subfig3}
\end{subfigure}%
\hfill
\begin{subfigure}[t]{0.23\textwidth}
\centering
\includegraphics[width=\textwidth, trim=0mm 5mm 0mm 6mm, clip]{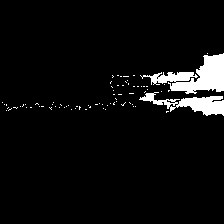}
\caption{Annotation area}
\label{fig:subfig4}
\end{subfigure}%
\caption{The figures depict an example of (a) OCT slice with corresponding (b) \textit{edge entropy} map,  (c) \textit{edge divergence} map, (d) query regions for annotation by our EdgeAL. The figures reveal that there is less visibility of retinal layer separation lines on the right side of the OCT slice, which could explain the model's high uncertainty in that region.}
\label{fig:four_images}
\end{figure}

Where $ D_{KL} \big ( P_{i}^{(m,n)} || \phi_{i}^{(m,n)} \big )$ measures the difference between model prediction probability and contextual probability for pixels belonging to edges of the input (Figure \ref{fig:subfig3}).
\label{selection}
\subsection{Superpixel Selection}
Clinical images have sparse representation, which can be beneficial for active learning annotation \cite{mackowiak2018cereals}. We use a traditional segmentation technique, SEEDS \cite{van2012seeds}, to leverage the local structure from images for finding superpixels. Annotating superpixels and regions for active learning may be more beneficial to the user than annotating the entire picture \cite{mackowiak2018cereals}.

We compute mean \textit{edge entropy} $EE_{i}^r$ and mean \textit{edge divergence} $ED_{i}^d$ for a particular area $r$ within a superpixel.%

\begin{equation}
   EE_{i}^{r} = \frac{1}{|r|} \sum_{(m,n) \in r} EE_{i}^{(m,n)}
   \label{eq_3}
\end{equation}

\begin{equation}
   ED_{i}^{r} = \frac{1}{|r|} \sum_{(m,n) \in r} ED_{i}^{(m,n)}
   \label{eq_4}
\end{equation}

Where $|r|$ is the amount of pixels in the superpixel region. 
We use regional entropy to find the optimal superpixel for our selection strategy and pick the one with the most significant value based on the literature \cite{siddiqui2020viewal}.
\begin{equation}
   ( i,r) = \underset{(j,s)}{\arg\max} \quad EE_{j}^{s}
\end{equation}
Following \cite{siddiqui2020viewal}, we find the subset of superpixels in the dataset with a 50\% overlap $(r, i)$. Let us call it set $R$. We choose the superpixels with the largest \textit{edge divergence} to determine the ultimate query (sample) for annotation.
\begin{equation}
   (p,q) = \underset{(j,s) \in R }{\arg\max} \big \{ ED_{j}^{s} \quad | \quad (j,s) \cap (i,r); (i,r) \in D_u)\}
\end{equation}

After each selection, we remove the superpixels from $R$. The selection process runs until we have $K$ amount of superpixels being selected from $R$. 

After getting the selected superpixel maps, we receive the matching ground truth information for the selected superpixel regions from the oracle. The model is then freshly trained on the updated labeled dataset for the next active learning iteration.

\section{Experiments and Results}
This section will provide a detailed overview of the datasets and architectures employed in our experiments. Subsequently, we will present the extensive experimental results and compare them with other state-of-the-art methods to showcase the effectiveness of our approach. We compare our AL method with nine well-known strategies: softmax margin \textbf{(MAR)} \cite{joshi2009multi}, softmax confidence \textbf{(CONF)} \cite{wang2016cost}, softmax entropy \textbf{(ENT)} \cite{luo2013latent}, MC dropout entropy \textbf{(MCDR)} \cite{gal2017deep}, Core-set selection \textbf{(CORESET)} \cite{sener2017active}, \textbf{(CEAL)} \cite{gorriz2017cost}, and regional MC dropout entropy \textbf{(RMCDR)} \cite{mackowiak2018cereals}, maximum representations \textbf{(MAXRPR)} \cite{yang2017suggestive}, and random selection \textbf{(Random)}. 
\subsection{Datasets and Networks}
To test EdgeAL, we ran experiments on Duke \cite{chiu2015kernel}, AROI \cite{9596934}, and UMN \cite{rashno2017fully} datasets in which experts annotated ground truth segmentations. Duke contains 100 B-scans from 10 patients, AROI contains 1136 B-scans from 24, and UMN contains 725 OCT B-scans from 29 patients. There are nine, eight, and two segmentation classes in Duke, AROI, and UMN, respectively. These classes cover fluid and retinal layers. Based on convention and dataset guidelines \cite{9596934,farshad2022net}, we use a 60:20:20 training: testing: validation ratio for the experiment without mixing one patient's data in any of the splits. Further, we resized all the images and ground truths to $224 \times 224$ using Bilinear approximation. Moreover, we run a 5-fold cross-validation (CV) on the Duke dataset without mixing individual patient data in each fold's training, testing, and validation set. Table \ref{tab:5foldcv} summarizes the 5-fold CV results.
\begin{table}[ht!]
\centering
\caption{The Table summarizes 5-fold cross-validation results (mean dice) for active learning methods and EdgeAL on the Duke dataset. EdgeAL outperforms other methods, achieving 99\% performance with just 12\% annotated data.}
\label{tab:5foldcv}
\begin{tabular}{|c|c|c|c|c|c|c|c|c|}
\hline
GT(\%) & RMCDR & CEAL &  CORESET & \textbf{EdgeAL} & MAR & MAXRPR  \\ \hline
2\%       & \textbf{0.40 \textpm 0.05} & \textbf{ 0.40 \textpm 0.05} &  0.38  \textpm 0.04  &  \textbf{0.40 \textpm 0.05}  & \textbf{0.40 \textpm 0.09} & \textbf{0.41 \textpm 0.04}   \\
12\%      & 0.44 \textpm 0.04 & 0.54 \textpm 0.04 & 0.44 \textpm 0.05   & \textbf{0.82 \textpm 0.03}  & 0.44 \textpm 0.03 & 0.54  \textpm 0.09  \\
22\%      & 0.63 \textpm 0.05 & 0.54 \textpm 0.04 &  0.62 \textpm 0.04   & \textbf{0.83 \textpm 0.03}  & 0.58 \textpm 0.04  & 0.67 \textpm 0.07    \\
33\%      & 0.58 \textpm 0.07 &0.55 \textpm 0.06 & 0.57 \textpm 0.04   & \textbf{0.81 \textpm 0.04}   & 0.67 \textpm 0.03 & 0.61  \textpm 0.03     \\
43\%      & 0.70 \textpm 0.03 & 0.79 \textpm 0.03 &0.69 \textpm 0.03   & \textbf{0.83 \textpm 0.02}  & 0.70 \textpm 0.04 & 0.80 \textpm 0.04   \\
\hline
100\%      & 0.82 \textpm 0.03 & 0.82 \textpm0.03  &  0.82 \textpm 0.03  & 0.82 \textpm 0.02  & 0.83 \textpm 0.02 & 0.83 \textpm 0.02 \\\hline
\end{tabular}
\end{table}

\begin{table}[ht!]
\centering
\caption{The table summarizes the test performance (mean dice) of various active learning algorithms on different deep learning architectures, including pre-trained weights, trained on only 12\% actively selected data from the Duke dataset. The results (mean \textpm sd) are averaged after running two times in two random seeds. Superscript 'r' represents ResNet, 'm' represents MobileNet version 3 backbones, and '\textdagger' indicates that the networks are initialized with pre-trained weights from ImageNet \cite{deng2009imagenet}.}
\label{tab:12p_result}
\begin{tabular}{|c|c|c|c|c|c|c|}
\hline
Arch.   & p100 & \textbf{EdgeAL} & CEAL & CORESET & RMCDR & MAXRPR\\ \hline
YN\textsuperscript{*}\cite{farshad2022net}    & \textbf{ 0.83 \textpm 0.02}  & \textbf{0.83 \textpm 0.01}   &  0.52 \textpm 0.01  & 0.45 \textpm 0.02    & 0.44  \textpm 0.01 &  0.56  \textpm 0.01\\
YN \cite{farshad2022net}  & 0.82 \textpm 0.02 & \textbf{0.81 \textpm 0.02}   & 0.48 \textpm 0.01  & 0.47  \textpm 0.02   & 0.45 \textpm 0.01 & 0.53  \textpm 0.01 \\ 
UN\cite{khan2020survey} & 0.79 \textpm 0.02 & \textbf{0.80 \textpm 0.01}   & 0.39 \textpm 0.01 & 0.48  \textpm 0.02   & 0.63 \textpm 0.01 & 0.51  \textpm 0.01 \\ 
DP-V3\textsuperscript{r} & 0.74 \textpm 0.04  & \textbf{0.74 \textpm 0.02}  & 0.62 \textpm 0.01  & 0.49  \textpm 0.01  & 0.57 \textpm 0.01  & 0.61  \textpm 0.01 \\
DP-V3\textsuperscript{m}  & 0.61 \textpm 0.01  & \textbf{0.61 \textpm 0.01}  & 0.28 \textpm 0.02  & 0.25  \textpm 0.01   & 0.59  \textpm 0.02 &  0.51  \textpm 0.01 \\ 
DP-V3\textsuperscript{r,\textdagger}   & 0.78 \textpm 0.01  & \textbf{0.79 \textpm 0.01}    & 0.29 \textpm 0.01 & 0.68  \textpm 0.01   & 0.68 \textpm 0.01 &  0.73  \textpm 0.01 \\ 
DP-V3\textsuperscript{m,\textdagger} & 0.78 \textpm 0.01  & \textbf{0.79  \textpm 0.01}  & 0.18 \textpm 0.01  & 0.57  \textpm 0.01   &  \textbf{ 0.79 \textpm 0.02}  & 0.75  \textpm 0.02\\ \hline
\end{tabular}
\end{table}
\begin{figure}
\includegraphics[trim=0 2mm 0 2mm,clip,width=1\linewidth]{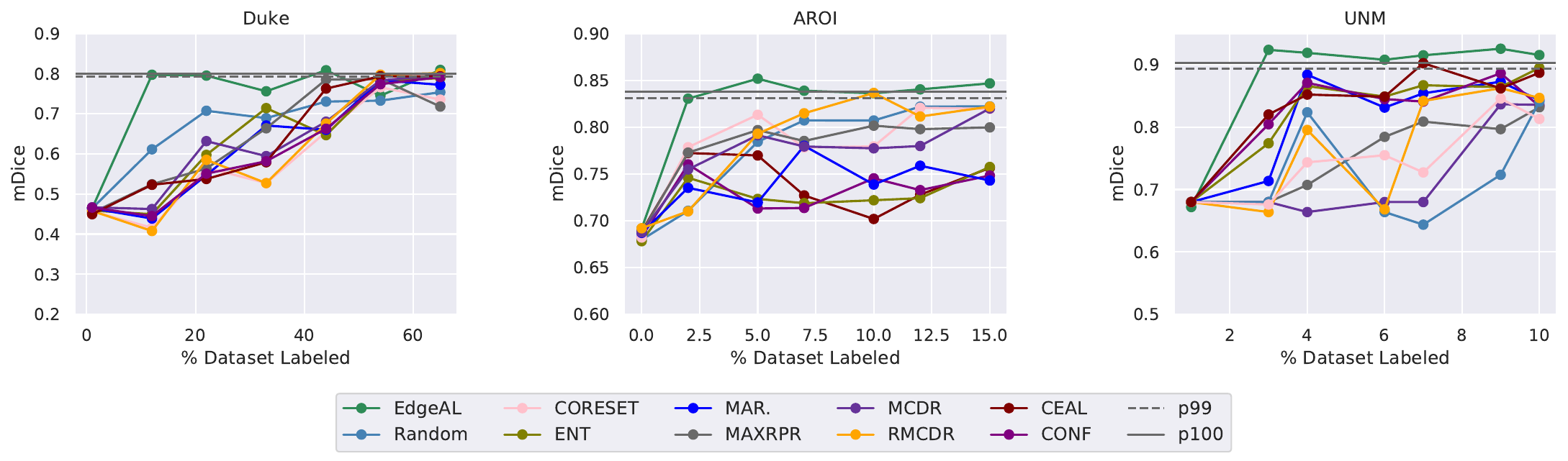}
\caption{EdgeAL's and other AL methods' performances (mean dice score) compared to baselines for Duke, AROI, and UNM datasets. Solid and dashed lines represent model performance and 99\% of it with 100\% labeled data.}
\label{fig2}
\end{figure}

We run experiments using Y-net(YN) \cite{farshad2022net}, U-net (UN) \cite{khan2020survey}, and DeepLab-V3 (DP-V3) \cite{siddiqui2020viewal} with ResNet and MobileNet backbones \cite{khan2020survey}. We present the results in Table \ref{tab:12p_result}.
No pre-trained weights were employed in the execution of our studies other than the ablation study presented in Table  \ref{tab:12p_result}. We apply mixed loss of dice and cross-entropy and  Adam as an optimizer, with learning rates of 0.005 and weight decay of 0.0004, trained for 100 epochs with a maximum batch size of 10 across all AL iterations. We follow the hyperparameter settings and evaluation metric (dice score) of \cite{farshad2022net}, which is the baseline of our experiment. 

\subsection{Comparisons}

Figure \ref{fig2} compares the performance of EdgeAL with other contemporary active learning algorithms across three datasets. Results show EdgeAL outperforms other methods on all 3 datasets. Our method can achieve 99\% of maximum model performance consistently with about 12\% ($\sim$8 samples), 2.3\% ($\sim$16 samples), and 3\% ($\sim$14 samples) labeled data on Duke, AROI, and UNM datasets. Other AL methods, CEAL, RMCDR, CORESET, and MAR, do not perform consistently in all three datasets. We used the same segmentation network YN\textsuperscript{*} and hyperparameters (described in Section \ref{sec3_1}) for a fair comparison. %

Our 5-fold CV result in Table \ref{tab:5foldcv} also concludes similarly. We see that after training on a 2\% seed set, all methods have similar CV performance; however, after the first active selection at 12\% training data, EdgeAL reaches close to the performance of full data training while outperforming all other active learning approaches.

Furthermore, to scrutinize if EdgeAL is independent of network architecture and weight initialization, we run experiments on four network architectures with default weight initialization of PyTorch (LeCun initialization)\footnote[1]{https://pytorch.org} and image-net weight initialization. Table \ref{tab:12p_result} presents the test performance after training on 12\% of actively selected data. These results also conclude that EdgeAL's performance is independent of the architecture and weight choices, while other active learning methods (RMCDR, MAXRPR) only perform well in pre-trained models.
\begin{figure}
\includegraphics[trim=0 2mm 0 2mm,clip,width=1\linewidth]{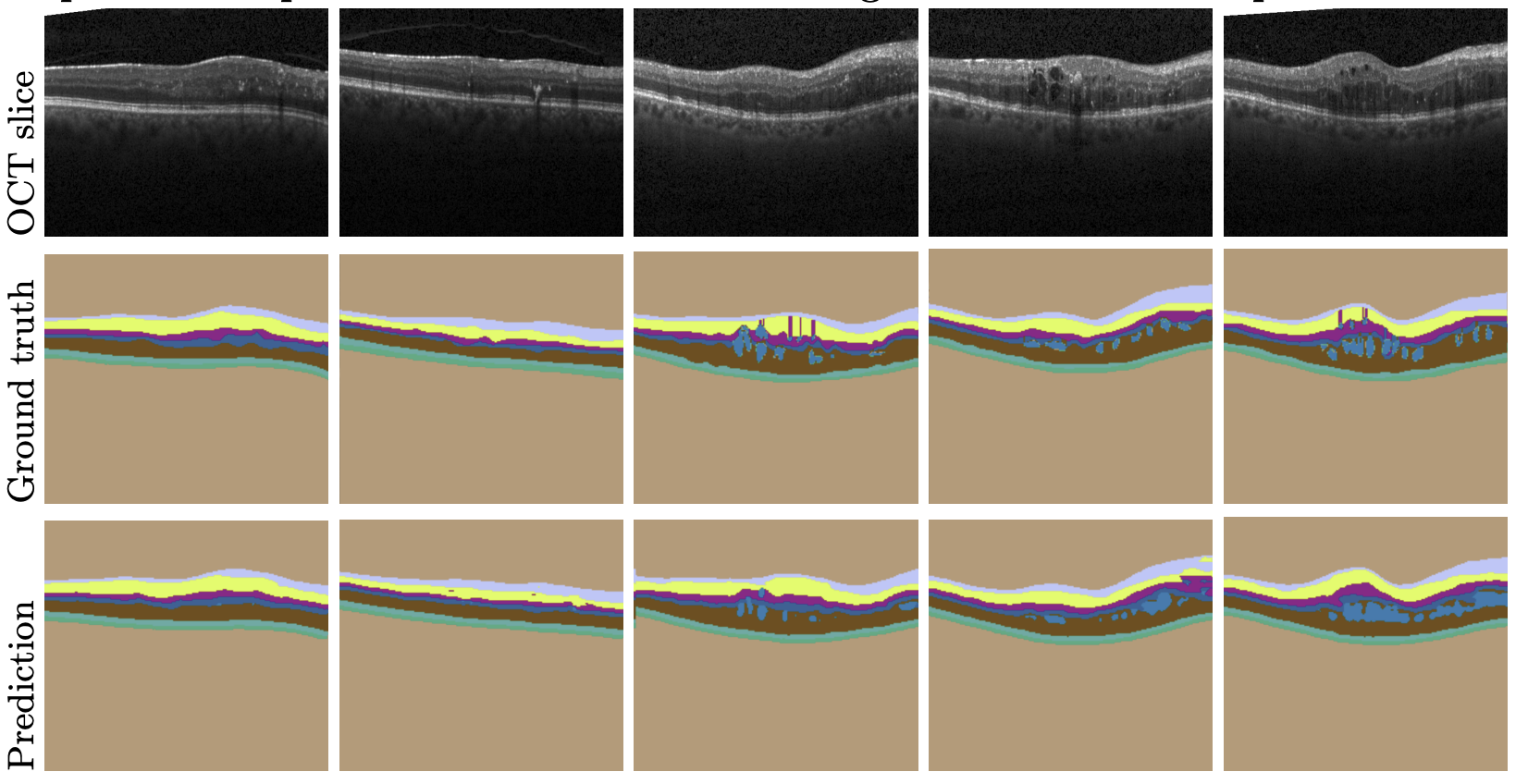}
\caption{Figures show sample OCT (Duke) test images with human-annotated ground truth segmentation maps and our prediction results, trained on just 12\% of the samples.}
\label{tab:sample_pred}
\end{figure}

\section{Conclusion}
EdgeAL is a novel active learning technique for OCT image segmentation, which can accomplish results similar to full training with a small amount of data by utilizing edge information to identify regions of uncertainty.
Our method can reduce the labeling effort by requiring only a portion of an image to annotate and is particularly advantageous in the medical field, where labeled data can be scarce. EdgeAL's success in OCT segmentation suggests that a significant amount of data is not always required to learn data distribution in medical imaging.
Edges are a fundamental image characteristic, allowing EdgeAL to be adapted for other domains without significant modifications, which leads us to future works.

\subsubsection{Acknowledgements} This work was partially funded by the German Federal Ministry of Education and Research (BMBF) under grant number 16SV8639 (Ophthalmo-AI) and supported by the Lower Saxony
Ministry of Science and Culture and the Endowed Chair of Applied Artificial Intelligence (AAI)
of the University of Oldenburg.
\bibliographystyle{splncs04nat}
\bibliography{references}
\end{document}